\documentclass{article}
\usepackage{xcolor}         
\usepackage{times}
\usepackage{latexsym}
\usepackage[T1]{fontenc}    
\usepackage[utf8]{inputenc} 
\usepackage{microtype}      
\usepackage{inconsolata}    
\usepackage[colorlinks,urlcolor=blue,linkcolor=blue,citecolor=blue]{hyperref}
\usepackage{url}
\usepackage{booktabs}
\usepackage{amsfonts}
\usepackage{nicefrac}
\usepackage{graphicx}
\usepackage{subfigure}
\usepackage{multirow}
\usepackage{adjustbox}
\usepackage{bigdelim}
\usepackage{comment}
\usepackage{tikz,pgfplots}
\pgfplotsset{compat=1.18}
\usepackage{colortbl}

\newcommand{\tablerowcolor}{\rowcolor[gray]{0.925}}
\newcommand{\tablecellcolor}{\cellcolor[gray]{0.925}}

\definecolor{ultramarine}{rgb}{0.07, 0.04, 0.56} 
\definecolor{blue}{rgb}{0.0, 0.4, 0.8} 
\definecolor{red}{rgb}{0.9,0.17,0.31} 
\definecolor{yellow}{rgb}{0.90196, 0.62549, 0.13922}  
\definecolor{green}{rgb}{0.00932, 0.49020, 0.19608} 
\definecolor{purple}{rgb}{0.4, 0.0, 0.8} 

\renewcommand{\baselinestretch}{1}

\title{Command-line Risk Classification using Transformer-based Neural Architectures}

\author{
    {\normalsize Paolo Notaro$^{1,2}$, Soroush Haeri$^{1}$, Jorge Cardoso$^{1,3}$, Michael Gerndt$^{2}$} \\
    \\
    \footnotesize $^1$Huawei Munich Research Center, Munich, 80992, Germany \\
    \footnotesize $^2$Technical University of Munich, Garching b. München, 85748, Germany \\
    \footnotesize $^3$CISUC, Department of Informatics Engineering, University of Coimbra, Coimbra, Portugal
}
\date{}

\begin{document}

\maketitle

\begin{abstract}
    To protect large-scale computing environments necessary to meet increasing computing demand, cloud providers have implemented security measures to monitor Operations and Maintenance (O\&M) activities and therefore prevent data loss and service interruption. Command interception systems are used to intercept, assess, and block dangerous Command-line Interface (CLI) commands before they can cause damage. Traditional solutions for command risk assessment include rule-based systems, which require expert knowledge and constant human revision to account for unseen commands. To overcome these limitations, several end-to-end learning systems have been proposed to classify CLI commands. These systems, however, have several other limitations, including the adoption of general-purpose text classifiers, which may not adapt to the language characteristics of scripting languages such as Bash or PowerShell, and may not recognize dangerous commands in the presence of an unbalanced class distribution. In this paper, we propose a transformer-based command risk classification system, which leverages the generalization power of Large Language Models (LLM) to provide accurate classification and the ability to identify rare dangerous commands effectively, by exploiting the power of transfer learning. We verify the effectiveness of our approach on a realistic dataset of production commands and show how to apply our model for other security-related tasks, such as dangerous command interception and auditing of existing rule-based systems.
\end{abstract}

\setcounter{secnumdepth}{0}

\maketitle

\section{Introduction}
The growth of security threats has encouraged organizations to develop effective security solutions to safeguard digital applications, data and resources~\cite{lee_apparatus_2016,microsoft,protected-shell}. Solutions target the prevention of incidents and malicious attacks. In large IT systems, Operations and Maintenance (O\&M) personnel accesses remote production systems to configure and repair services. O\&M operators execute thousands of operations each day, and incorrect or malicious commands can cause major system failures and data losses resulting in high monetary and reputation costs.

Security solutions are required to prevent any service impact or infrastructural damage by recognizing potentially dangerous operations before they are executed. A typical solution is to mediate the access to production systems through a bastion host~\cite{safe-proxies}.

\begin{table}[t]
    \centering
    \begin{adjustbox}{width=0.75\linewidth,center}
    \begin{tabular}{c c c}
    \hline
        \textbf{Command} & \textbf{Risk Class} & \textbf{Notes} \\
         \cmidrule[0.4pt](lr{0.125em}){1-1}%
         \cmidrule[0.4pt](lr{0.125em}){2-2}%
         \cmidrule[0.4pt](lr{0.125em}){3-3}%
        \texttt{rm 2023-04-21\_12:45:67.log} & LOW & deletes temporary log\\ \tablerowcolor
        \texttt{rm -rf /bin/*} & HIGH & (force) deletes executables dir\\ 
        \texttt{cat \$DELETE\_LIST | grep *.log} & LOW & prints filtered content\\ \tablerowcolor
        \texttt{cat \$DELETE\_LIST | xargs -0 rm} & HIGH & deletes files from input file\\
        \texttt{time ls} & LOW & (timed) \texttt{cwd} directory listing\\ \tablerowcolor
        \texttt{time kill -9 12345} & HIGH & process kill hidden by \texttt{time} call\\ 
        \texttt{echo `kill 7890'} & LOW & prints a string \\ \tablerowcolor
        \texttt{echo \`{}kill 7890\`{}} & HIGH & backtick evaluates \texttt{kill}\\ \hline
    \end{tabular}
    \end{adjustbox}
\vspace{3mm}
\caption{Examples of commands with varying complexity and risk. It can be easily observed how small changes to parameters, flags, and some uses of the command-line syntax highly influence the risk of executed commands.}
\label{tab:command-complexity}
\end{table}

The evaluation of command risk is an open and challenging problem, due to large variability in executable programs and their combination of arguments, flags, and environment variables (see Table~\ref{tab:command-complexity}). Rule-based classifiers are simple and frequently adopted solutions for this problem - however, in large-scale environments, they require to regular maintenance and revision of existing rules to reflect the dynamism of the environment.

Previous works in Natural Language Processing (NLP) have shown the high capabilities provided by machine learning approaches, and specifically Large Language Models (LLMs), on a variety of NLP-related tasks~\cite{bert-survey}. Their double learning procedure, based on pretraining the model on a large-scale, domain-specific corpus of text, and finetuning the model for the specific task with supervised data, allows to provide higher context understanding and generalization power than previous NLP algorithms.

Some previous works in the context of O\&M security~\cite{microsoft,leibniz} have applied machine learning to command-line data for security-related classification tasks, such as malicious command detection. However, no previous approach has considered the potentiality of LLMs for command classification, or it has tried to adapt existing LLM-powered solutions for text classification for the CLI security domain.

To this end, we propose the use of a transfer-learning, transformer-based method for the command risk classification task. Our main contribution is a language model pretrained on large corpus of command-line language data, which can be adapted for several language-related tasks via finetuning. Using a curated dataset of real-world production commands, we applied our language model for command risk classification, where we can obtain risk predictions that are more accurate and more sensitive towards potentially dangerous commands. 
Being our model trained end-to-end to process and classify commands, it does not rely on manual verification from humans during runtime and it provides 100\% classification, differently from rule-based systems. We evaluate and compare our approach to existing algorithms for command risk classification. 
Results show that our approach improves detection of dangerous commands, and it is particularly able to pinpoint highly dangerous commands more effectively (F1-score +22\%), albeit such commands are extremely rare and thus limited in the training data. We also describe how to apply our pretrained command-line language model for other security-related tasks.

The rest of the paper is organized as follows. In Section~\nameref{sec:background}, we present foundations on command-line security 
and related work. In Section~\nameref{sec:approach}, the proposed approach is presented, followed by~\nameref{sec:experimental-setup} and experimental \nameref{sec:results}. In Section~\nameref{sec:use-cases}, we discuss  for ourthe applicability of our approach in the context of cloud security. 
We summarize the outcome of the paper in the~\nameref{sec:conclusion}.


\section{Background}
\label{sec:background}
\vspace{5mm}

In this section we present the foundations of command-line security and the command risk classification problem (\nameref{ssec:cli-security}) 
 and a summary of related work (\nameref{ssec:related-work}).

\subsection{Command-line Security}
\label{ssec:cli-security}
\vspace{3mm}

\begin{figure}[t]
    \centering
    \resizebox{\columnwidth}{!}{\includegraphics{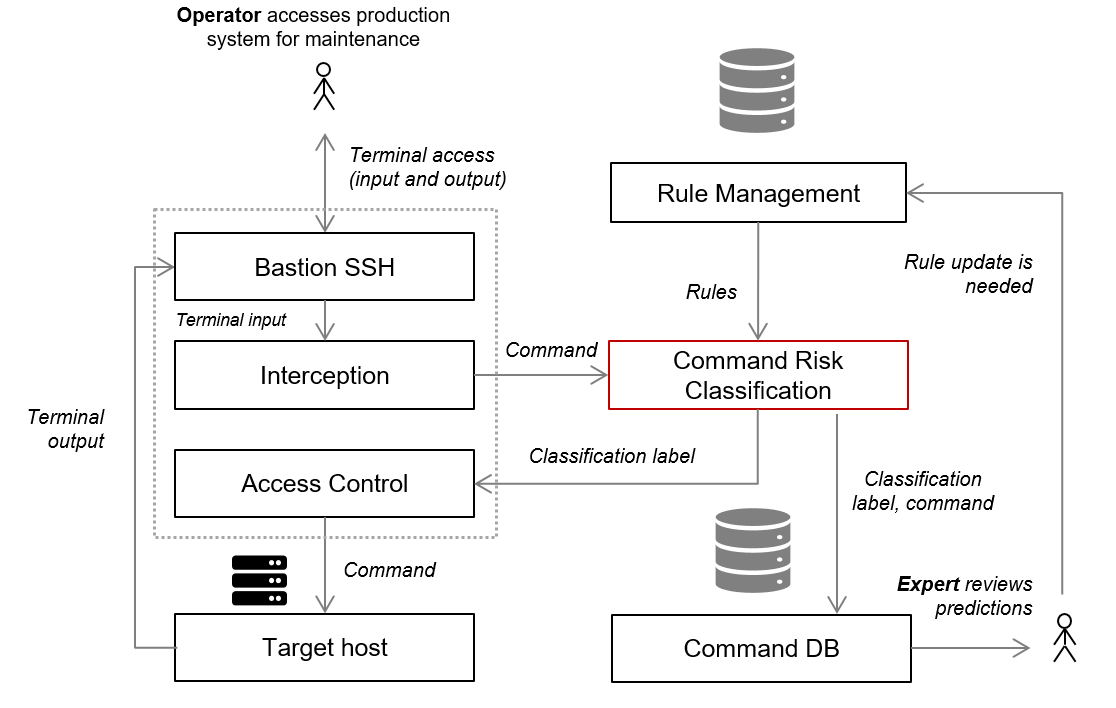}}
    \caption{Architecture of a rule-based risk assessment system. Commands executed by operators are intercepted by a bastion host (Bastion SSH) to be evaluated using a set of rules stored in a configuration DB (Rule Management). If the command is evaluated safe, it is forwarded to the Target Host, otherwise an error is reported. All risk evaluations are logged and periodically revised by security experts, who may update the rules.}
    \label{fig:cli-security-system}
    \vspace{-5mm}
\end{figure}

Large-scale computing environments require high levels of availability and reliability to fulfill Service Level Agreements (SLA). Due to the large scale of devices and services being executed, failures are, albeit rare, inevitable. To this end, Operation and Maintenance (O\&M) personnel operates to resolve and minimize the impact of failures on the correct runtime of computing services. 

O\&M requires operators to access remote systems to configure and repair services via Command-line Interface (CLI). 
Providing free access to remote production systems poses two major security challenges, namely 1) the possibility of external attackers gaining unauthorized access to systems and 
2) the possibility of operators accidentally performing harmful operations and damage service availability. 
For this reason, operators typically access remote hosts via a bastion host, which allows to review, approve, and execute commands without establishing direct connection to systems~\cite{safe-proxies}. Bastion hosts implement an interception system, which captures executed commands and analyses them before granting execution rights.


The need to recognize and block dangerous commands before their execution requires an algorithm to estimate command risk. Different risk classes may be defined, and each class may be associated with the necessary privilege to execute commands. Command risk classification is then the association of each incoming command to one of the predefined privilege and/or risk classes. Based on the classifier decision and the current user privileges, the bastion host allows or blocks the current command, and returns output the user.


A common solution for estimating command risk is a rule-based classifier, where IF-THEN-ELSE rules define which commands are allowed (or \textit{whitelisted}) and which commands are blocked (or \textit{blacklisted}).
The rules may be defined based on the expertise of O\&M operators and the historical records of executed commands, and stored in a configuration database that allows to periodically revise and update them. Rule-based systems may implement a specification syntax, based on regular expressions or quantifiers, to cover a larger set of commands with a single expression. 

Rule-based systems are simple, easily configurable and explainable. However, they also have several limitations: 
\begin{enumerate}
    \item new combination of programs and arguments may be executed, for which the existing rules are not suitable;
    \item the risk level assigned to rules by operators based on their expertise, may diverge from the true risk of commands
; 
    \item they require a default handling action when a command does not match any existing rule. This default action is however limiting, as a "default block" strategy may hinder important operations during incident response, while a "default allow" may allow dangerous operations to be executed;
    \item dealing with the complexity of command-line syntax is difficult without resorting to a complex pattern language for rules  (see Figure~\ref{tab:command-complexity}).
\end{enumerate}


Therefore, human supervision is required to make sure the rules reflect the risk of executed commands over time. A revision is performed by comparing commands with their corresponding predicted risk class. Since the majority of commands executed are harmless and dangerous commands are rarely being executed, the manual verification of commands is a tedious and error-prone job. The problem may be further aggravated by an inaccurate interception system, which may capture console content, such as command outputs, password prompts, auto-completions, etc. which does not constitute an executable input and must be ignored.

\subsection{Related Work}
\label{ssec:related-work}
\vspace{3mm}

Hendler et al.~\cite{microsoft} evaluate several machine learning models for malicious PowerShell command detection. Both traditional NLP (n-gram, BoW) and deep neural network models (CNN~\cite{zhang2016characterlevel}, LSTM) are considered, including an ensemble combining a 3-gram and a CNN model, which yields the best performance. During the preprocessing phase, characters are one-hot encoded, with case information at character-level provided as an input binary flag. The CNN model applies 1D convolutional layers based on the architecture proposed by Zhang et al.~\cite{zhang2016characterlevel} on input commands padded to fixed length. The use of character-level one-hot encoding with a closed vocabulary does not allow to effectively model the semantics and inter-relationship between input tokens. To this end, we choose to apply Byte-pair Encoding~\cite{bpe} and Wordpiece embeddings~\cite{wordpiece}, which enable the model to learn the most frequent tokens directly from the training data and associate co-occurring and meaning-related tokens in the embedding space.
Moreover, a fully-supervised model requires a large quantity of training data to achieve high accuracy. Our model can leverage information learned during pretraining for classification and thus only requires a limited dataset for finetuning to the context-specific tasks, providing more flexibility at a reduced effort.

Yamin et al.~\cite{yamin2019Windows} use Naive Bayes and CNN models on command-line arguments to classify PowerShell commands as malicious. They evaluated their classification accuracy on a dataset composed of 14 categories of commands. Their approach focuses on PowerShell and obfuscated command detection. Results are evaluated in terms of accuracy (96\%) and the dataset class distribution is not provided. Because dangerous commands are rare, the class distribution is typically highly unbalanced, which allows to construct trivial classifiers that can achieve high accuracy. For this reason, we focus on evaluating our model on the positive class (i.e. dangerous commands) by measuring precision and recall, to ensure our model can also classify these rare yet important commands. 

\begin{figure*}[t]
    \centering
    \resizebox{\linewidth}{!}{\includegraphics{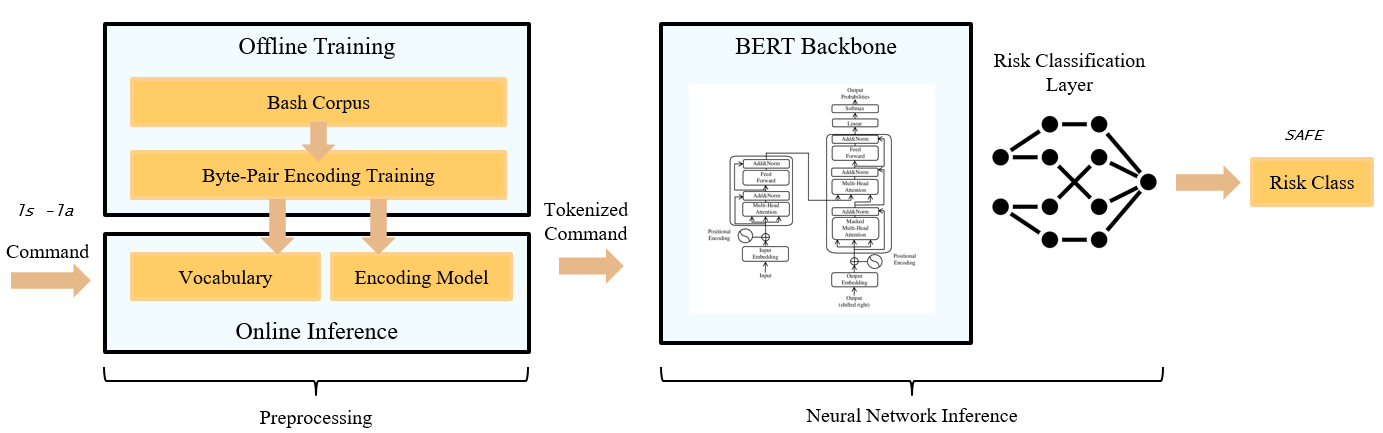}}
    \caption{AI Classifier architecture. The input command is preprocessed via Byte-Pair Encoding to construct an input sequence of tokens. The sequence is processed by the BERT backbone to produce a latent representation of the command, which encodes important language-related information learned during pretraining. This latent representation is given to the risk classification layer to estimate the final command risk.}
    \label{fig:system-architecture}
\end{figure*}

PyComm~\cite{pycomm} is a malicious command detection model for Python scripts. It is based on random forest applied on a hybrid set of static features and Python source code strings. During evaluation, they obtained an accuracy of 0.955 with a recall of 0.943. Differently from them, our approach focuses on command-line risk classification and rely exclusively on the command string.

The similarity of commands is estimated by processing the documentation of commands using NLP techniques.

\section{Approach}
\label{sec:approach}
\vspace{5mm}

In this section, we describe our approach for command risk classification. We describe the system architecture and the training procedure.

\subsection{System Architecture}
\label{ssec:architecture}
\vspace{3mm}

Our risk classification system is based on Bidirectional Encoder Representations from Transformers (BERT)~\cite{bert}. BERT is a deep language representation model based on transfer learning~\cite{bert}. BERT has empirically shown to improve performance in many language understanding tasks, including question answering, text classification, sentence pair completion, named-entity recognition, etc. ~\cite{bert, bert-survey, bert-classification}. In particular, BERT has shown to provide more effective results for discriminative tasks, compared to the effectiveness of other approaches (e.g. GPT3~\cite{gpt3}) for generative tasks. Therefore, we select BERT for our command risk classification task.

The key concept of BERT is to pretrain a transformer deep neural network on contextual tasks, such as masked token and next sentence prediction, to learn the language syntax and the contextual relationships between tokens present in the language. Because these contextual tasks are self-supervised learning tasks, the pretraining step in BERT does not require any labeled data, and a large corpus of the target language is sufficient. In a second phase, called finetuning, the pretrained BERT model can be adapted for the specific task to perform (e.g.,~text classification). In the finetuning phase, the BERT model learns the new task while retaining language knowledge from the pretraining step, enabling a higher level of generalization.

The complete system architecture is shown in Figure~\ref{fig:system-architecture}. Our approach is composed of a preprocessing algorithm and a neural network architecture, which is trained as described in Section~\nameref{ssec:training}. 

In the preprocessing step, the raw command string is split into a sequence of tokens using the Byte-Pair Encoding (BPE) algorithm~\cite{bpe, bpe-orig}. BPE is an unsupervised tokenization method, in which the most frequently occurring pair of characters is recursively replaced with a character that does not occur in the vocabulary. This allows to estimate the more frequent combination of characters and use it as tokens for the language. We select BPE over traditional space-based tokenization for two main reasons~\cite{intellicode}: 
\begin{itemize}
    \item BPE will identify the optimal set of tokens, i.e. the set of tokens that are more frequent in the observed language;
    \item The identified tokens will be composed of one or more characters, but they will be shorter than traditional words, which will make the model more robust towards out-of-vocabulary tokens. This increases the coverage provided by the tokenization system without resorting to a large fixed vocabulary;
\end{itemize}

The BPE-encoded tokens are fed into a bidirectional transformer architecture~\cite{attention}. The token indices are represented as Wordpiece embeddings~\cite{wordpiece} and then fed through the encoder-decoder architecture of the transformer. The transformer architecture is composed of $n_A$ attention heads~\cite{attention} of size $H$. To perform classification, we append a classification layer block with output size $C$ to the transformer output. The final output is a list of class probabilities associated to the different risk classes. The class with the highest associated probability is selected as the predicted class. 

\subsection{Training}
\label{ssec:training}
\vspace{3mm}

\begin{figure*}[t!]
    \centering     
    \resizebox{\linewidth}{!}{\includegraphics{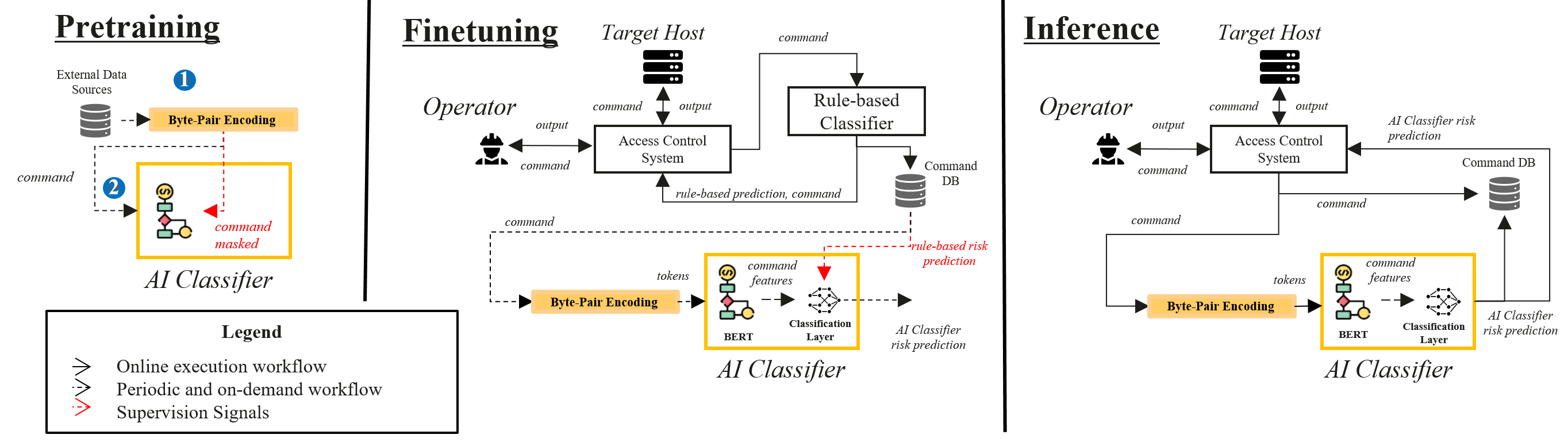}}
    \caption{System architecture during the three phases of construction of the AI classifier. During pretraining, a command corpus is used to learn the language tokens and their context relationships. During finetuning, a dataset of labeled commands is used to specialize the AI model for the risk classification task. Both commands and labels are originating from the interception system composed of a rule-based classifier. During inference, the AI classifier replaces the rule-based classifier providing online risk classification for all commands executed.}
    \label{fig:phases-diagram}
\end{figure*}

Figure~\ref{fig:phases-diagram} summarizes the construction phases of the AI classifier. Our training procedure is composed of three steps: dataset collection, BPE training, BERT training (pretraining and BERT finetuning).

\subsubsection{Dataset Collection.}
\label{sssec:dataset-collection}

For our pretraining phase, requiring large-quantity of raw command data, we programmatically collected a corpus of Bash files from publicly available data.  First, we searched for GitHub repositories with the Bash language tag. Then, from each of these repositories we selected only Bash-related files, by matching specific criteria (first line contains a shebang, or the file has a $.sh$ extension). This resulted in a final collection of 71164 Bash scripts, amounting to about 500 MB. 

\subsubsection{BPE Training.}
\label{sssec:bpe-training}

During this phase, our corpus of Bash commands is used to learn the tokens and patterns of the scripting language in use, using the BPE algorithm described above (\nameref{ssec:architecture}). The frequency of different character-level patterns is estimated based on the observed data, to produce a vocabulary of $V$ learned tokens of size and an encoding algorithm to extract them from raw commands string.

\subsubsection{BERT Training.}
\label{sssec:bert-training}

The BERT pretraining step requires to train the transformer network on self-supervised contextual tasks. Therefore, we first annotate each sample from our Bash corpus with corresponding task labels. After this annotation step, the total dataset size amounts to 15 GB. During the pretraining phase, the transformer model is pretrained on the contextual tasks described in ~\cite{bert}: bidirectional masked LM, by masking 15\% of sequence tokens at random; and next sentence prediction, i.e. by predicting the next command in the sequence of commands inside a Bash script, or a random command. After pretraining, only the network backbone, which outputs, a $h_{i}$-long representation of the input command is retained, while the contextual output layers are dropped.


During the finetuning phase, the pretrained transformer backbone is extended with an additional classification layer block to enable command risk classification. The classification layer block is composed of a FC linear layer preceded by a dropout layer and followed by a softmax normalization layer. For this training phase, a supervised dataset of commands, annotated with risk classes, is used. The network pipeline is trained end-to-end using the softmax loss, to maximize the log likelihood of the training dataset. The network weights are updated using gradient descent, with gradients computed via back-propagation.

\section{Experimental Setup}
\label{sec:experimental-setup}
\vspace{5mm}

\begin{table}[b!]\vspace*{-\baselineskip}
\scriptsize 
\centering
\begin{adjustbox}{width=0.75\linewidth,center}
    \begin{tabular}{ c c c c }
    \cmidrule[0.4pt](lr{0.125em}){2-4}%
        & \textbf{parameter} & \textbf{value} & \textbf{description} \\
        \cmidrule[0.4pt](lr{0.125em}){2-2}%
        \cmidrule[0.4pt](lr{0.125em}){3-3}%
        \cmidrule[0.4pt](lr{0.125em}){4-4}%
        \ldelim\{{7}{*}[{\rotatebox[origin=c]{90}{\scriptsize NN architecture}}] &$H$ & $256$ & hidden size\\ 
        & $p_{d}$ & $0.1$ & dropout probability \\
        & $n_A$ & $4$ & attention heads \\ 
        & $n_h$ & $4$ & hidden layers \\
        & $h_i$ & $1024$ & intermediate size\\ 
        &$V$ & $20000$ & vocabulary size\\
        & $L$ & $1024$ & max sequence length\\ 
        \ldelim\{{5}{*}[{\rotatebox[origin=c]{90}{\scriptsize NN training}}] & $\sigma_{init}$    & $0.02$ & initializer range\\
        & $B$ & $128$ & batch size\\ 
        & $E$ & $16$ & epochs \\
        & $C$ & $3$ & output classes \\ 
        & $\epsilon$ & $3\cdot10^{-4}$ & learning rate \\ 
        \ldelim\{{4}{*}[{\rotatebox[origin=c]{90}{\scriptsize ML training}}] & $c$ & $10^2$ & LR inverse regularize factor \\ 
        & $S$ & $50$ & Word2Vec $H$ \\
        & $\alpha$ & $0.05 \rightarrow 0.0007$ & Word2Vec $\epsilon$ (start/min) \\ 
        & $E$ & $100$ & Word2Vec $E$ \\ \cmidrule[0.4pt](lr{0.125em}){2-4}%
        
    \end{tabular}
    \end{adjustbox}
    \caption{Hyper-parameter configuration used in our command risk classification experiments.}
    \label{tab:hyperparameters}
\end{table}

In this section, we describe the experimental setup used in our training experiments, as well as for the evaluation of all prediction models (\nameref{ssec:evaluation-metrics}).

We implemented our approach and all evaluated models in Python. We selected the \textit{4/256 BERT mini} model as our architecture. All network layers utilize GELU activation functions and are trained using the Adam optimizer. The full list of numerical hyperparameters is shown in Table~\ref{tab:hyperparameters}. We pretrained our transformer model for 350,000 iterations (or 160 epochs). The total training time required 10 days on a single NVIDIA v100 GPU.

In addition to our pretraining corpus described in~\nameref{sssec:dataset-collection}, we have collected a second dataset of commands annotated with class risk. This dataset is used for training and evaluating our classification model and all models presented in~\nameref{ssec:evaluation-metrics}. This dataset contains a realistic collection of commands used for O\&M. 

As our annotated commands originate from our internal cloud production systems, invalid samples resulting from errors in the interception system, such as password prompts, non-Bash commands, and bash terminal output have been automatically filtered out and then manually verified. Misspelled and unknown commands have been preserved, as they still represent executable commands and may cause damage. Multi-program commands (resulting from pipelining, \texttt{xargs}, ...) and script file calls have also been preserved, as they are representative of real commands executed in production.  

Both commands and corresponding risk labels are retrieved from an internal data store, used for offline analysis of risk predictions as described in~\nameref{sec:background} and as depicted in Figure~\ref{fig:phases-diagram}. The risk labels originate from a rule-based system, and have been manually verified by experts and corrected in case of discrepancies with true command risk. In our scenario, we consider three risk classes:
\begin{itemize}
    \item SAFE, for read-only commands and commands that do not alter the state of the system significantly;
    \item RISKY, for commands that may irreversibly alter the state of the system and cause damage, for which privilege escalation is required;
    \item BLOCKED, for commands that will irreversibly alter the correct state of the system, and must never be executed.
    
\end{itemize}


To ensure our annotated dataset is representative of a real command distribution, we studied the distribution of commands in the three classes described above. After removing invalid commands, we have observed an approximate 80\%, 20\% split between SAFE and RISKY commands, with an additional 0.3\% component of (extremely rare) BLOCKED commands.

Therefore, after expert verification of the risk labels, we down-sampled our annotated dataset to replicate the class ratios observed in our class analysis. In the end, we collect 47158 high-quality commands, representative of a real O\&M workload. We divided our finetuning dataset into \textit{train}, \textit{dev}, and \textit{test} splits with 70\%, 20\%, 10\% ratios, while preserving the class distribution. 

\subsection{Evaluation Metrics}
\label{ssec:evaluation-metrics}
\vspace{3mm}

\begin{table*}[tb]
    \centering
    \begin{adjustbox}{width=\linewidth,center}
    \begin{tabular}{c c c c c c c c c c c}
    \cmidrule[0.4pt](lr{0.125em}){2-11}%
        & \multirow{2}{*}{\textbf{Model}} & \multicolumn{3}{c}{\textbf{Precision}} & \multicolumn{3}{c}{\textbf{Recall}} & \multicolumn{3}{c}{\textbf{F1-score}} \\ \cmidrule[0.4pt](lr{0.125em}){3-5} %
        \cmidrule[0.4pt](lr{0.125em}){6-8} %
        \cmidrule[0.4pt](lr{0.125em}){9-11}
        & & \textbf{RISKY} & \textbf{BLOCKED} & \textbf{R+B} & \textbf{RISKY} & \textbf{BLOCKED} & \textbf{R+B} & \textbf{RISKY} & \textbf{BLOCKED} & \textbf{R+B} \\ \cmidrule[0.4pt](lr{0.125em}){2-11}
        & \tablecellcolor Word2Vec & \tablecellcolor 0.9210 & \tablecellcolor 1.0000 & \tablecellcolor 0.9440 & \tablecellcolor 0.6658 & \tablecellcolor 0.2857 & \tablecellcolor 0.6638 & \tablecellcolor 0.7729 & \tablecellcolor 0.4444 & \tablecellcolor 0.7795\\ 
        \ldelim\{{6}{*}[{\rotatebox[origin=c]{90}{\small \cite{microsoft}}}] & 3-gram & 0.9322 & 0.8571 & 0.9328 & 0.8576 & 0.7500 & 0.8578 & 0.8934 & 0.8000 & 0.8937\\
        & \tablecellcolor BoW & \tablecellcolor 0.7018 & \tablecellcolor 0.3212 & \tablecellcolor 0.7048 & \tablecellcolor 0.3212 & \tablecellcolor 0.3750 & \tablecellcolor 0.3228 & \tablecellcolor 0.4407 & \tablecellcolor 0.5000 & \tablecellcolor 0.4428\\ 
        & 4-cnn & 0.9391 & \textbf{1.0000} & 0.9420 & 0.8894 & 0.6250 & 0.8893 & 0.9136 & 0.7692 & 0.9149\\
        & \tablecellcolor 9-cnn & \tablecellcolor 0.9420 & \tablecellcolor 0.3077 & \tablecellcolor 0.9430 & \tablecellcolor 0.8212 & \tablecellcolor 0.5000 & \tablecellcolor 0.8287 & \tablecellcolor 0.8774 & \tablecellcolor 0.3810 & \tablecellcolor 0.8821\\
        & LSTM & 0.9451 & - & 0.9507 & 0.7894 & 0.0000 & 0.7867 & 0.8603 & - & 0.8610\\
        & \tablecellcolor DTEnsemble & \tablecellcolor 0.9600 & \tablecellcolor 0.8333 & \tablecellcolor 0.9603 & \tablecellcolor 0.9035 & \tablecellcolor 0.6250 & \tablecellcolor 0.9021 & \tablecellcolor 0.9309 & \tablecellcolor 0.7143 & \tablecellcolor 0.9303\\ 
        & \textit{ours} & \textit{\textbf{0.9713}} & \textit{\textbf{1.0000}} & \textit{\textbf{0.9716}} & \textit{\textbf{0.9165}} & \textit{\textbf{0.8750}} & \textit{\textbf{0.9161}} & \textit{\textbf{0.9431}} & \textit{\textbf{0.9333}} & \textit{\textbf{0.9430}}\\ \cmidrule[0.4pt](lr{0.125em}){2-11}
    \end{tabular}
    \end{adjustbox}
    \caption{Evaluation Results in terms of precision, recall and F1-score on the positive classes RISKY and BLOCKED, for different algorithms. (R+B) = all dangerous commands (RISKY + BLOCKED). Bold value indicates the best model for each metric. "-" indicates the metric cannot be computed due to no TP predictions.}
    \label{tab:accuracy-results}
\end{table*} 

We evaluated our risk classification model in terms of accuracy, precision, recall, and F1-score~\cite{salfner-fp-survey}, for their ability to measure predictor ability also in the presence of imbalanced class distributions. We measure these metrics the two positive classes of our dataset, as we are focusing on the detection of dangerous commands, so we do not evaluate the SAFE class metrics. We compare our approach to:
\begin{itemize}
    \item a baseline model, constructed using Word2Vec embedding ~\cite{word2vec} + Random Forest algorithm;
    \item the NLP and neural-based approaches presented in~\cite{microsoft}, namely a 3-gram model, a BoW model, two one-dimensional CNNs of 4 and 9 layers (called \textit{4-cnn} and \textit{9-cnn}, respectively), a LSTM-based neural network, and the ensemble model combining the \textit{3-gram} and \textit{4-cnn} models.
\end{itemize}
    
We reproduce the implementations of these algorithms to the best of our knowledge based on the information available, re-adopting the original hyper-parameters when provided.

\section{Results}
\label{sec:results}
\vspace{5mm}

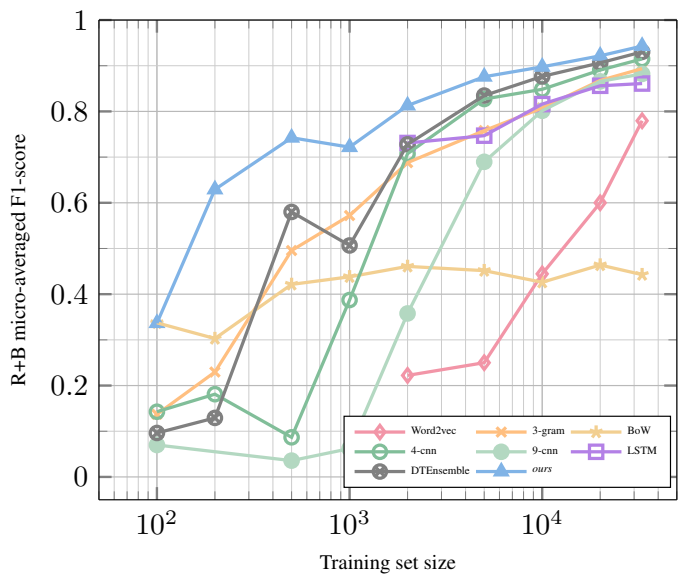
\begin{figure}[b!]
\centering
\resizebox{0.75\columnwidth}{!}{
\begin{tikzpicture}%
\begin{axis}[%
    set layers, mark layer=axis tick labels,
    separate axis lines,
    every outer x axis line/.append style={black},
    xmajorgrids,
    xminorgrids,
    ymajorgrids,
    grid=both,
    every minor tick/.append style={ultra thin},
    every axis/.append style={thin, tick style={thick}},
    every axis plot post/.append style={line width=1.1pt, mark size=2.25pt},
    minor grid style={gray!40, ultra thin},
    major grid style={gray!55, thin},
    label style ={font=\scriptsize},
    xlabel shift = 0.15,
    ylabel shift = 0.15,
    minor y tick num={1},
    minor x tick num={1},
    scatter/classes={a={mark=o,draw=black}},
    xlabel=Training set size,
    ylabel=R+B micro-averaged F1-score,
    xmode=log,
    xmin = 50, %
    xmax = 50000, %
    ymin = -0.05, %
    ymax = 1, %
    legend columns=3, %
    legend style={nodes={scale=0.75, transform shape},at={(0.99, 0.1)}, anchor=east,font=\tiny,
    legend cell align=left, align=left}
]
\addplot[color=red!50,mark=diamond, mark size=3] table [scatter,scatter src=explicit symbolic, x=training set size, y=word2vec+RF, col sep=comma] {res/eval-data-size.csv};
\addplot[color=orange!50,mark=x, mark size=3] table [scatter,scatter src=explicit symbolic, x=training set size, y=3-gram, col sep=comma] {res/eval-data-size.csv};
\addplot[color=yellow!50,mark=star, mark size=3] table [scatter,scatter src=explicit symbolic, x=training set size, y=bow, col sep=comma] {res/eval-data-size.csv};
\addplot[color=green!50,mark=o, mark size=3] table [scatter,scatter src=explicit symbolic, x=training set size, y=4-cnn, col sep=comma] {res/eval-data-size.csv};
\addplot[color=green!30,mark=*, mark size=3] table [scatter,scatter src=explicit symbolic, x=training set size, y=9-cnn, col sep=comma] {res/eval-data-size.csv};
\addplot[color=purple!50,mark=square, mark size=3] table [scatter,scatter src=explicit symbolic, x=training set size, y=LSTM, col sep=comma] {res/eval-data-size.csv};
\addplot[color=black!50, mark=otimes, mark size=3] table [scatter,scatter src=explicit symbolic, x=training set size, y=DTEnsemble, col sep=comma] {res/eval-data-size.csv};
\addplot[color=blue!50, mark=triangle*, mark size=3] table [scatter,scatter src=explicit symbolic, x=training set size, y=ours, col sep=comma] {res/eval-data-size.csv};
    
\legend{Word2vec,3-gram,BoW,4-cnn,9-cnn,
LSTM,DTEnsemble,\textit{ours}};

\end{axis}
\end{tikzpicture}}%
\caption{F1-score of RISKY + BLOCKED commands on test set, as a function of dataset size used for training. We can observe how our BERT approach can classify dangerous commands more accurately in the presence of limited training data. Missing points indicate the F1-score could not be computed due to no TP predictions.}
\label{fig:performance-size}
\vspace{-5mm}
\end{figure}

Table~\ref{tab:accuracy-results} presents the comparison of the different models under evaluation, in terms of precision, recall, and F1-score on the two positive classes RISKY and BLOCKED.
Our approach achieves the highest absolute score for 8 of 9 metrics measured. Over the second-best result, the precision in detecting RISKY commands is improved by 1.13\%, while the recall is increased by 1.30\%. For BLOCKED commands, we measured an increase of 16.7\% in precision and 25.0\% recall. On average, for all dangerous commands, our approach can improve precision by 1.13\% and recall by 1.40\%. The increase in F1-score for all dangerous commands is 1.27\%. If we consider the ratio of dangerous commands (20\%) and we assume an average number of 3M commands/month executed in production, which is consistent with our experience, such increase in recall results in approximately 60k additional dangerous commands intercepted during operations, which may cause an equivalent number of potential incidents.

To show the transfer learning abilities of our model, we also evaluated the ability of predictors to recognize dangerous commands with limited training data. We randomly sampled 100, 200, 500, 1000, ..., 20000 commands from our training set and trained the corresponding models. Results are reported in Figure~\ref{fig:performance-size}. It can be observed how our model achieves the best F1 performance on dangerous commands for all reduced dataset sizes, while requiring one order of magnitude less samples to achieve comparable F1-score results.

\section{Use Cases}
\label{sec:use-cases}
\vspace{5mm}

The BERT model we trained with BPE can be applied in many industrial applications related to the command-line, which is the entrance door to the cloud.
We here describe several use cases for our model that we have used or plan to use in the future.


The first use case we discuss is online risk classification. Command risk can be evaluated online to block dangerous commands during interception. 

A scheme of online risk classification is shown in the last segment of Figure~\ref{fig:phases-diagram}. First, commands executed over remote terminal are intercepted by an access control system (e.g.~a bastion host). Then, the risk of the intercepted command is evaluated using our LLM-based classifier. The classifier can be deployed directly on-site, or accessed via an inference API to take advantage of specialized hardware. Commands and classifier predictions can be stored to permanent storage for offline analysis. Based on the prediction outcome, commands are either blocked or allowed, and the operation output displayed to the operator.

System auditing is the practice of analyzing the quality of an existing system to validate its results and consider potential improvements. In the context of command interception systems, an existing risk classifier can be audited by analyzing if its risk predictions correspond to the true risks of the commands, to support the identification of errors and the creation of new classification rules. 

We applied our model for auditing an existing rule-based system. An example of our auditing pipeline is shown in Figure~\ref{fig:cli-security-system}. The existing system is composed of a rule-based risk classifier, a rule management system, and database store of rule-based and AI-based predictions. When revision of existing rules is needed, a report of non-matching of predictions from the rule-based and AI models is generated. As for the majority of commands the two predictions will correspond, it is sufficient to report only the commands where the two predictions differ. An human expert can then decide if the discrepancy reported by the AI model is correct and update the corresponding rule in the management system. This comparison speeds up the work of expert reviewers, as they only need a small portion of commands, where the two predictions do not match. Thanks to the high precision of our classifier, if a command is reported as dangerous, it is likely that an rule update action must be taken. 
Using our model, we were able to discover several new risky commands that were not detected by the rule-based system.


We have also considered our transformer model for command categorization. Commands can be assigned to predefined categories based on their function (e.g.~networking, filesystem, scripting, third-party command). For O\&M operators it is convenient to know the category of a command for several reasons: to identify similar commands and construct new rules, to clarify the meaning of unknown commands and speed up auditing, to understand which type of commands are currently not identified by the security system. The command categorization problem is comparable to command risk classification, as both can be achieved using the techniques described in this work (provided ground-truth labels are available). 
In our experiments, we considered categories from manpages documents~\cite{manpages} as potential source of command categorization labels. 
First, commands are parsed to extract program names. Then, program names are matched to manpage files, which store a corresponding command categories (e.g. \texttt{ls =`OS command'}). This allows to create a labeled dataset associating commands with one (or more) command categories. By applying the finetuning technique described in Section~\nameref{sssec:bert-training}, it is possible to construct a language-specific command classifier for an arbitrary command taxonomy.

We also considered our approach for other NLP-related tasks, such as context extraction, where it can be used to automatically extract code from mixed-language text, e.g. code tutorials and Jupyter Notebooks, and from Standard Operating Procedures (SOPs) for O\&M practices. The extraction of commands from SOP allows to verify the correct execution of the procedure and potentially recognize SOPs and commands that do not conform to the correct security standards.
Our approach can also be used for named-entity recognition tasks in a command-line context, e.g. to recognize filenames, API endpoints, or IP addresses. This task can support security auditing by identifying named entities that should not be accessed.








\section{Conclusion}
\label{sec:conclusion}
\vspace{5mm}

In this paper, we proposed a language model for the command-line language, which is applicable to several NLP-related tasks. We showed how to apply our approach for command risk classification. Our language model leverages the contextual knowledge learned during pretraining to achieve higher classification accuracy and pinpoint dangerous commands effectively. We described the procedure to train our model according to a realistic distribution of production commands. We evaluated the accuracy of our approach and compared it to a several existing approaches for command risk classification. Our results show that it can improve detection for rare classes of commands. 


\def\namerefname{REFERENCES}
\bibliography{custom.bib}
\bibliographystyle{IEEEtran}

\end{document}